\documentclass{l4dc2025}

\title[Disentangling Uncertainties]{Disentangling Uncertainties by Learning Compressed Data Representation}
\usepackage{times}
\usepackage{algorithm}
\usepackage{algorithmic}
\usepackage{enumitem}
\usepackage{booktabs}
\usepackage{url}
\usepackage{wrapfig}
\usepackage{titlesec}
\usepackage{multirow}
\usepackage[bf]{caption} 

\titlespacing*{\section}
{0pt}{0.5em}{0.2em}
\titlespacing*{\subsection}
{0pt}{0.3em}{0.1em}

\author{%
 \Name{Zhiyu An} \Email{zan7@ucmerced.edu}\\
 \addr University of California, Merced
 \AND
 \Name{Zhibo Hou} \Email{zhou6@ucmerced.edu}\\
 \addr University of California, Merced
 \AND
 \Name{Wan Du} \Email{wdu3@ucmerced.edu}\\
 \addr University of California, Merced
}

\begin{document}

\maketitle

\begin{abstract}%

We study aleatoric and epistemic uncertainty estimation in a learned regressive system dynamics model. Disentangling aleatoric uncertainty (the inherent randomness of the system) from epistemic uncertainty (the lack of data) is crucial for downstream tasks such as risk-aware control and reinforcement learning, efficient exploration, and robust policy transfer. While existing approaches like Gaussian Processes, Bayesian networks, and model ensembles are widely adopted, they suffer from either high computational complexity or inaccurate uncertainty estimation. To address these limitations, we propose the Compressed Data Representation Model (CDRM), a framework that learns a neural network encoding of the data distribution and enables direct sampling from the output distribution. Our approach incorporates a novel inference procedure based on Langevin dynamics sampling, allowing CDRM to predict arbitrary output distributions rather than being constrained to a Gaussian prior. Theoretical analysis provides the conditions where CDRM achieves better memory and computational complexity compared to bin-based compression methods. Empirical evaluations show that CDRM demonstrates a superior capability to identify aleatoric and epistemic uncertainties separately, achieving AUROCs of 0.8876 and 0.9981 on a single test set containing a mixture of both uncertainties. Qualitative results further show that CDRM's capability extends to datasets with multimodal output distributions, a challenging scenario where existing methods consistently fail. Code and supplementary materials are available at \url{https://github.com/ryeii/CDRM}.

\end{abstract}


\section{Introduction}

The disentanglement of epistemic and aleatoric uncertainties plays a crucial role in safe control and reinforcement learning \cite{brunke2022safe}. One notable example is the problem of efficient exploration with limited resources. Consider a robot operating with a finite power supply for world exploration. To optimize its power usage, the robot must selectively choose actions that generate the most informative data for reducing its world model's uncertainty. This process is known as curiosity-driven exploration \cite{pathak2017curiosity}. However, model uncertainty can stem from two distinct sources: \textit{epistemic uncertainty}, which arises from a lack of data (e.g., the unknown furniture arrangement in an unexplored room) and is reducible through additional data collection; and \textit{aleatoric uncertainty}, which results from inherent environmental randomness (e.g., the outcome of a dice throw) and remains irreducible regardless of additional data (see \cite{hullermeier2021aleatoric} for a review). This distinction becomes critical for resource-efficient exploration, as it would be wasteful for the robot to expend its limited power gathering data in hopes of predicting an inherently random event. States with irreducible aleatoric uncertainty are called \textit{stochastic traps} \cite{shyam2019model} or \textit{noisy TVs} \cite{schmidhuber1991adaptive}, with the latter term derived from the observation that a naively curious agent could dwell on the unpredictability of a noisy TV screen.

Several existing works aim to avoid the noisy TV problem through uncertainty estimation using Deep Ensemble \cite{lakshminarayanan2017simple, burda2018exploration, pathak2019self}, Gaussian Process (GP) \cite{wachi2018safe}, or learning a separate model to predict uncertainty \cite{mavor2022stay}, but they either cannot accurately disentangle the two types of uncertainties or have high computational complexity. Aside from efficient exploration, uncertainty estimation is also important for ensuring the safety of reinforcement learning agents \cite{brunke2022safe}. For example, the variance output by a GP system dynamics model is used as an estimation of epistemic uncertainty, which is used to filter potentially unsafe actions \cite{an2023clue, zhao2023probabilistic}. These works also suffer from inaccurate estimation and high computational complexity.

Traditional uncertainty estimation for regressive models has strong baselines such as GP \cite{duvenaud2014automatic}, MC Dropout \cite{gal2016dropout}, Bayesian Network \cite{blundell2015weight}, and Deep Ensemble \cite{lakshminarayanan2017simple}. However, they suffer from the aforementioned disadvantages. Deep Evidential Learning (DER) \cite{amini2020deep} uses neural networks to predict the parameters of the output distribution. To get an accurate prediction, it requires the output distribution type to be known, while in real-world datasets such distribution types are often unknown. \cite{chan2024hyper} proposes a hyper-network that produces the parameters of multiple downstream diffusion networks and uses the downstream networks as an ensemble for prediction. This method shows empirical improvement compared to Deep Ensemble but suffers from higher computational complexity due to more model forward passes. On the other hand, density-based methods \cite{sun2023flagged, yoon2024uncertainty, bui2024density} have shown promising results in out-of-distribution (OOD) detection or novelty detection for classification tasks. These methods estimate the data density of the dataset at a given input; the data density is then used to decide whether the input is OOD. Such methods have not yet been extensively adopted and tested in regressive tasks.

\begin{figure}
    \captionsetup{format=plain}
    \centering
    \includegraphics[width=\linewidth]{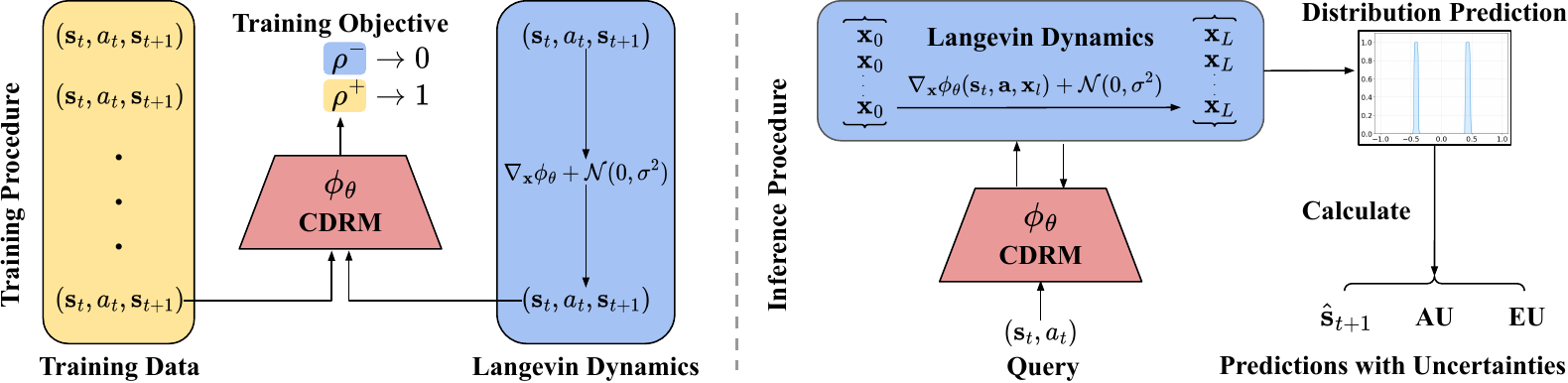}
    \vspace{-1.5em}
    \caption{An overview of the training (left) and inference (right) procedures of CDRM. The training procedure simultaneously generate positive and negative samples by sampling the training data and running Langevin dynamics sampling, respectively. The losses of positive and negative samples are combined and propagated through CDRM, and the procedure is repeated. The inference procedure runs Langevin dynamics sampling on the CDRM output in the subset of the input space with the given state-action pair, and output a distribution of CDRM's output across a range of next state. This distribution is then used to calculate AU, EU, and next state prediction.}
    \vspace{-1.5em}
    \label{fig: overall}
\end{figure}

In this paper, we explore applying data distribution representation learning to improve uncertainty estimation for a system dynamics regression task. We follow recent work on Energy-Based Models (EBM) \cite{du2020improved} for their efficient training process, using Langevin dynamics sampling to train a neural network that contains a compressed representation of the dataset, named \textit{Compressed Data Representation Model }(\textbf{CDRM}). The input for CDRM is a (state, action, next-state) tuple, and the output is a scalar indicating if this tuple is present in the dataset. Through this process, CDRM learns a compressed representation of the dataset. We then designed an efficient inference procedure using Langevin sampling to query the CDRM and obtain estimated next-state distributions for given (state, action) pairs. The learned data representation enables CDRM to sample from arbitrary next-state distributions, rather than being constrained to a Gaussian prior. Empirically, we observe that CDRM accurately estimates the next-state distribution even when it has multiple modes. While multimodal next-state distributions do not conform to the assumption of a single underlying dynamics function, a multimodal distribution more accurately captures the true distribution of inherently random real-world events, such as a dice throw, while keeping the state-space continuous. An overview of the training and inference procedures of CDRM is shown in Figure \ref{fig: overall}. Our contributions are summarized as follows:

\begin{itemize}[leftmargin=*]
\itemsep 0.3em 
  \item We propose Compressed Data Representation Model (CDRM), a framework to estimate epistemic and aleatoric uncertainty for a system dynamics regression task using a learned compressed data representation.
  \item We design an efficient inference procedure using Langevin dynamics sampling to query CDRM for prediction and uncertainty estimates.
  \item We provide theoretical analysis on the memory and computational complexities of CDRM and show the condition where CDRM surpasses bin-based compression.
  \item We benchmark CDRM with competitive baseline methods on dedicated dataset for disentangling aleatoric and epistemic uncertainties. Quantitative and qualitative results demonstrated CDRM's advantage at uncertainty disentanglement and multimodal distribution prediction.
\end{itemize}

\section{Preliminaries}

\subsection{Problem Definition} \label{Sec. problem definition}

Given system dynamics $\mathbf{s}_{t+1} = f(\mathbf{s}_t, \mathbf{a})$ where $\mathbf{s}_t$ is the current state, $\mathbf{a}$ is the action, and $\mathbf{s}_{t+1}$ is the next state, we consider estimating aleatoric and epistemic uncertainties of a given input $(\mathbf{s}_t, \mathbf{a})$ using a transition dataset $\mathcal{D}:\{(\mathbf{s}_t, \mathbf{a}, \mathbf{s}_{t+1})\}_N$. Here, we describe the definition of uncertainties: 

\noindent\textbf{Aleatoric Uncertainty (AU)} arises from the inherent randomness of a system that is irreducible by additional data. Existing works \cite{chan2024hyper} often model AU as a Gaussian distribution, with a real-world example being a sensor experiencing irreducible noise signals in the form of Gaussian noise added to the true signal. However, we note that many events with inherent randomness do not conform to this formulation. In the example of the noisy TV problem \cite{schmidhuber1991adaptive}, the state distribution of the TV signal can be an arbitrary distribution. Without knowing the state distribution prior, assuming a Gaussian distribution can lead to misleading estimates. In this paper, we explore a procedure to estimate arbitrary next-state distributions.

\noindent\textbf{Epistemic Uncertainty (EU)} arises from a lack of knowledge, i.e., the absence of data about a given input. EU can also arise from poorly fitted models due to insufficient training \cite{hullermeier2021aleatoric}. This type of uncertainty is reducible with additional data or stronger convergence through additional training epochs. Unlike AU, EU is dependent on the available training data and hence cannot be faithfully represented by a model which has a loss function defined on a distribution over model parameters \cite{bengs2022pitfalls}. In this paper, we measure EU using posterior distribution of model parameters.

\subsection{Maximum Likelihood Training with Binary Labels}

Training a model to distinguish \textit{positive} and \textit{negative} samples is a simple yet useful technique. It was employed to train Energy-Based Models \cite{du2020improved} where the data distribution is trained to have low energy, and other region is trained to have high energy. More recently, it is used to fine-tune reinforcement learning policies \cite{hejna2023contrastive} and large language models \cite{rafailov2024direct}. The common idea shared by these works is that, given an input $\mathbf{x}$, positive output $\mathbf{y}^+$, and negative output $\mathbf{y}^-$, the maximum likelihood training aims to minimize the loss:
\vspace{-0.5em}
\begin{equation} 
L = \mathbb{E}\left[\log P(\mathbf{y}^+|\mathbf{x}) - \log P(\mathbf{y}^-|\mathbf{x})\right]
\vspace{-0.5em}
\end{equation}
where $P(\mathbf{y}^+|\mathbf{x})$ represents the likelihood of output being $\mathbf{y}^+$ given input $\mathbf{x}$. For a model parameterized by $\theta$, the gradient of the loss is found by $\nabla_\theta L = \mathbb{E}\left[\nabla_\theta \log P(\mathbf{y}^+|\mathbf{x}) - \nabla_\theta \log P(\mathbf{y}^-|\mathbf{x})\right]$. Intuitively, maximum likelihood training pushes the output landscape of model $\mathcal{M}_\theta$ at the position $\mathbf{x}$ towards $\mathbf{y}^+$ and away from $\mathbf{y}^-$.

\subsection{Maximum Likelihood Training with Langevin Dynamics Sampling} \label{Sec. Maximum Likelihood Training with Langevin Dynamics Sampling}

While the basic maximum likelihood training with binary labels provides a direct way to shape the model's output distribution, it requires explicit negative samples $\mathbf{y}^-$. In many cases, generating appropriate negative samples can be challenging or computationally expensive.
Let $y = \mathcal{M}_\theta(\mathbf{x})$ be the model output given $\mathbf{x}$, consider the problem of minimizing $\mathcal{M}_\theta(\mathbf{x})$ for $\mathbf{x}\in\mathcal{D}$ and maximizing $\mathcal{M}_\theta(\mathbf{x})$ otherwise. 
To sample $\arg\min_{\mathbf{x}}\mathcal{M}_\theta(\mathbf{x})$ efficiently, \cite{du2019implicit, du2020improved} employed Langevin dynamics sampling as a Markov Chain Monte Carlo (MCMC) procedure that uses stochastic gradient descent as the transition kernel. The sampling process is defined as:
\vspace{-0.5em}
\begin{equation}
\mathbf{x}_{t+1} = \mathbf{x}_t - \epsilon\nabla_\mathbf{x}\mathcal{M}_\theta(\mathbf{x}) + \sqrt{2\epsilon}\mathcal{N}(0, \mathbf{I})
\vspace{-0.5em}
\end{equation}
where $\epsilon$ is the step size and $t$ is the iteration index. After sufficient iterations, $\mathbf{x}_T$ approximates a solution to $\arg\min_{\mathbf{x}}\mathcal{M}_\theta(\mathbf{x})$. The training objective can then be written as:
\vspace{-0.5em}
\begin{equation}
L = \mathbb{E}_{\mathbf{x}\sim\mathcal{D}}\left[\mathcal{M}_\theta(\mathbf{x})\right] - \mathbb{E}_{\mathbf{x}\sim\mathcal{X}\setminus\mathcal{D}}\left[\mathcal{M}_\theta(\mathbf{x})\right]
\vspace{-0.5em}
\end{equation}
In \cite{du2020improved}, $\mathcal{M}_\theta$ is defined by an Energy-Based Model \cite{lecun2006tutorial}. Langevin dynamics sampling has several advantages: (i) it automatically generates diverse negative samples that reflect the model's current beliefs; (ii) the sampling process adapts as the model learns, maintaining effective training signals; (iii) the exploration enabled by noise helps prevent mode collapse.

\section{Learning and Inferring from Compressed Data Representation}

Consider modeling system dynamics $\mathbf{s}_{t+1} = f(\mathbf{s}_t, \mathbf{a})$. While the classical approach involves gathering transition datasets $\mathcal{D}:\{(\mathbf{s}_t, \mathbf{a}, \mathbf{s}_{t+1})\}_N$ and learning a regressive model $\hat{f}_\theta$ parameterized by $\theta$ to approximate true dynamics $f$, we propose an alternative approach. We construct CDRM $\phi_\theta:\mathcal{S}\times\mathcal{A}\times\mathcal{S}\rightarrow [0, 1]$, a deep neural network parameterized by weights $\theta$ that maps a transition data point to a real scalar in $[0, 1]$. CDRM consists of two components: a deep neural network (DNN) and the sigmoid function $(1+\exp(x))^{-1}$ to map the output of the DNN to the desired range.
\vspace{-0.5em}
\begin{equation}\label{Eq. CDRM}
    \phi_\theta(\mathbf{s}_t, \mathbf{a}, \mathbf{s}_{t+1}) = \text{Sigmoid}\circ\text{DNN}_\theta(\mathbf{s}_t, \mathbf{a}, \mathbf{s}_{t+1}) = \begin{cases}
        \rho^+ & \text{if } (\mathbf{s}_t, \mathbf{a}, \mathbf{s}_{t+1}) \in \mathcal{D} \\
        \rho^- & \text{otherwise}
    \end{cases}
\end{equation}
The DNN architecture can be either a multi-layer perceptron (for processing tabular-format inputs) or convolutional layers (for acoustic or image inputs), provided the final layer outputs a scalar. We use $\rho$ to denote the output of the CDRM. During training on dataset $\mathcal{D}$, if the input transition data point $(\mathbf{s}_t, \mathbf{a}, \mathbf{s}_{t+1})$ belongs to $\mathcal{D}$, it is termed a \textit{positive sample} with output denoted as $\rho^+$; otherwise, the input is a \textit{negative sample} with output denoted as $\rho^-$. 

The training objective is to enable the CDRM to distinguish between positive and negative samples. Specifically, we aim to maximize $\mathbb{E}(\rho^+) - \mathbb{E}(\rho^-)$, encouraging $\rho^+$ to approach 1 and $\rho^-$ to approach 0. This is achieved through the contrastive loss $L = -\mathbb{E}[\log(\rho^+ + \epsilon)] - \mathbb{E}[\log(1 - \rho^- + \epsilon)]$, where $\epsilon=1e-6$ ensures numerical stability. For model training procedure, CDRM follows the maximum likelihood training with Langevin dynamics sampling discussed in Section \ref{Sec. Maximum Likelihood Training with Langevin Dynamics Sampling}.

\subsection{Inference Algorithm}

For a given state-action pair $(\mathbf{s}_t, \mathbf{a})$, the inference algorithm determines three values: the predicted next state $\hat{\mathbf{s}}_{t+1}$, the epistemic uncertainty (EU), and the aleatoric uncertainty (AU). This requires identifying next states $\mathbf{s}_{t+1}$ where the model output $\rho = \phi_\theta(\mathbf{s}_t, \mathbf{a}, \mathbf{s}_{t+1})$ approaches 1, indicating valid transitions present in dataset $\mathcal{D}$. We formalize this notion through a threshold parameter $\alpha \in (0, 1)$, where $\rho > \alpha$ indicates sufficient model confidence.

Let $\mathbf{x}$ denote a candidate next state prediction $\hat{\mathbf{s}}_{t+1}$. A candidate is considered \textit{valid} if $\phi_\theta(\mathbf{s}_t, \mathbf{a}, \mathbf{x}) > \alpha$. This formulation transforms our task into searching a multi-dimensional, potentially non-convex objective function $F(\mathbf{x}) = \phi_\theta(\mathbf{s}_t, \mathbf{a}, \mathbf{x})$ to find $\mathbf{x}$ where $F(\mathbf{x})>\alpha$. While Markov Chain Monte Carlo (MCMC) methods are commonly employed for such tasks, two significant challenges emerge. \textbf{(i) High dimensionality:} Despite MCMC's advantages over standard Monte Carlo methods, it remains susceptible to the curse of dimensionality, particularly with high-dimensional states such as images \cite{du2020improved}. \textbf{(ii) Aleatoric uncertainty:} When aleatoric uncertainty exists in the dataset for a state-action pair $\mathbf{s}_t, \mathbf{a}$, multiple valid next states $\mathbf{s}_i, \mathbf{s}_j$ exist where $(\mathbf{s}_t, \mathbf{a}, \mathbf{s}_i), (\mathbf{s}_t, \mathbf{a}, \mathbf{s}_j) \in \mathcal{D}$ and $\mathbf{s}_i \neq \mathbf{s}_j$. Consequently, quantifying AU requires the set of valid candidates $\mathcal{X} = \{\mathbf{x}|\phi_\theta(\mathbf{s}_t, \mathbf{a}, \mathbf{x}) > \alpha\}$ rather than a single solution.

To address these challenges, we employ an efficient batch-parallel Langevin dynamics sampling approach. The procedure begins by initializing $N$ candidates sampled uniformly from the state space, processed simultaneously on GPU architecture. Each candidate is updated for $L$ steps using Langevin dynamics, combining gradient ascent with Gaussian noise to escape local optima. Invalid states are projected back to the state space boundary through efficient parallel clipping operations.

\subsection{Uncertainty Quantification}

We propose the following prediction and uncertainty quantification equations that aims to accurately capture the data distribution:
\vspace{-0.5em}
\begin{equation}
    \text{EU} = \begin{cases}
        1 & \text{if }\mathcal{X} = \varnothing \\
        (\text{KDE}_{\text{base}} + (1-\max_{\mathbf{x}\in\mathcal{X}}\phi_\theta)\cdot\sigma_{\phi})/2 & \text{otherwise}
    \end{cases}, \quad
    \text{AU} = \sqrt{\text{Var}\left[\mathcal{X}\right]}
    \vspace{-0.5em}
\end{equation}

where $\text{KDE}_{\text{base}}$ is an EU estimate of the input based on the Kernel Density Estimation (KDE) of this input given the dataset. This EU estimate is given by $\left(1 + \exp((\text{KDE}_{(\mathbf{s}_t, \mathbf{a})} - \mu_{\text{KDE}})/\sigma_{\text{KDE}})\right)^{-1}$, where $\mu_{\text{KDE}}, \sigma_{\text{KDE}}$ are the mean and standard deviation of the KDE of all points in the dataset pre-computed before inference. For our implementation, we used radial basis function as the density estimation kernel. $\sigma_{\phi}$ is the standard deviation of densities across Langevin steps. This EU equation combines two key sources of epistemic uncertainty: the distance-based density from the dataset and the CDRM's predictive confidence. $\text{KDE}_{\text{base}}$ captures OOD uncertainty by comparing the input's density estimation to the training distribution through sigmoid normalization. The second term $(1-\max{\phi})\cdot\sigma_{\phi}$ measures the model's prediction uncertainty, where $1-\max{\phi}$ reflects the confidence in the highest-density prediction and $\sigma_{\phi}$ captures the stability of density estimates across Langevin steps. We take the average of two EU estimates to reflect both sources of EU.

The next state prediction is given by $\hat{\mathbf{s}}_{t+1} = \arg\max_{\mathbf{x}\in\mathcal{X}}\phi_\theta(\mathbf{s}_t, \mathbf{a}, \mathbf{x})$. We deliberately avoid using the average or weighted average of valid samples like in Deep Ensemble \cite{lakshminarayanan2017simple} or BNN \cite{blundell2015weight}. This is because we do not assume the training data to be unimodal, i.e., the output data distribution for any input can be a multimodal distribution. In such cases, it is easy to see that the result of averaging two modes can fall outside the distribution, resulting in unrealistic predictions. Therefore, CDRM outputs the valid sample with the highest $\rho$ value, ensuring the prediction fits the data distribution. Finally, we define AU as the variance of all samples in the valid sample set.
The complete inference algorithm is summarized in Algorithm \ref{Alg. CDRM Inference}.

\begin{algorithm}
\algsetup{linenosize=\tiny}
\scriptsize
\caption{CDRM Inference}\label{Alg. CDRM Inference}
    \begin{algorithmic}[1]
        \REQUIRE CDRM $\phi_\theta$, Input $(\mathbf{s}_t, \mathbf{a})$, Sample No. $N$, Langevin steps $L$, Langevin step size $\beta$, Noise scale $\sigma$, Threshold $\alpha$, $\mu_{\text{KDE}}, \sigma_{\text{KDE}}$.
        \STATE $\mathcal{X}\gets\varnothing;\;\;\{\mathbf{x}_0\}_N \sim \mathcal{U}(\mathcal{S})$ \quad$\rhd$ Parallel sample initialization
        \FOR{$l = 1$ to $L$}
            \STATE $\mathbf{x}_{l} \gets \mathbf{x}_{l-1} + \beta\nabla_\mathbf{x}\phi_\theta(\mathbf{s}_t, \mathbf{a}, \mathbf{x}_{l-1}) + \mathcal{N}(0, \sigma^2)$ \quad$\rhd$ Batch Langevin update
            \STATE $\mathbf{x}_{l} \gets \text{clip}(\mathbf{x}_{l}, \mathcal{S})$ \quad$\rhd$ Parallel state space projection
            \STATE $\mathcal{X} \gets \mathcal{X}\cup\{\mathbf{x}_{l}|\mathbf{x}_{l}\notin\mathcal{X}, \phi_\theta(\mathbf{s}_t, \mathbf{a}, \mathbf{x}_{l}) > \alpha\}$
        \ENDFOR
        \IF{$\mathcal{X} \neq \varnothing$}
            \STATE Compute weights $w(\mathbf{x})$ for all $\mathbf{x}\in\mathcal{X}$
            \STATE $\hat{\mathbf{s}}_{t+1}\gets \arg\max_{\mathbf{x}\in\mathcal{X}}\phi_\theta(\mathbf{s}_t, \mathbf{a}, \mathbf{x})$,\quad EU $\gets (\text{KDE}_{\text{base}} + (1-\max_{\mathbf{x}\in\mathcal{X}}\phi_\theta)\cdot\sigma_{\phi})/2$, \quad AU $\gets \sqrt{\text{Var}\left[\mathcal{X}\right]}$
        \ELSE
            \STATE $\hat{\mathbf{s}}_{t+1}\gets$ None;\quad EU $\gets 1$;\quad AU $\gets$ None
        \ENDIF
        \RETURN $\hat{\mathbf{s}}_{t+1}$, EU, AU
    \end{algorithmic}
\end{algorithm}

\subsection{Implementation Details and Control Knobs}

In practice, the output of DNN in Equation \ref{Eq. CDRM} may have a large absolute value. This causes gradient vanish of the sigmoid function because $\sigma'(x) = \sigma(x)(1-\sigma(x)) \rightarrow 0$ as $|x|\rightarrow 0$. Vanishing gradient prevents Langevin dynamics sampling from effectively explore the model landscape. To address this, we clip the output of DNN to $[\epsilon, 1-\epsilon]$ where $\epsilon$ is small (e.g., $10^{-6}$). This ensures the minimum gradient magnitude is $\epsilon(1-\epsilon)\nabla_{\mathbf{x}}\text{DNN}_\theta(\mathbf{s}_t, \mathbf{a}, \mathbf{x})$, maintaining effective exploration during Langevin sampling while preserving the model's discriminative capability between positive and negative samples. The control knobs include hyper-parameters Langevin sample size $S$, number of Langevin steps $L$, valid threshold $\alpha$, Langevin step size $\beta$, and Langevin noise scale $\sigma$.

\section{Theoretical Analysis}

We provide theoretical analysis to show the practicality of CDRM. For small-scale problem with low dimension state spaces, uncertainty estimation through naive bin-based method could have better computational complexity than CDRM. A bin search-based method compresses data by discretizing state and action spaces into a number of bins. A flag vector is then constructed, such that a binary flag is represents the existence of data in each bin. We show that for problems requiring high granularity or having high dimensional state space, CDRM has lower computational complexity and significantly lower memory complexity than bin-based method.
\vspace{-0.3em}
\begin{remark}\label{Remark inference bin-search-based method}
    Consider a dataset $\mathcal{D}$ containing tuples $(\mathbf{s}_t, \mathbf{a}, \mathbf{s}_{t+1})$, where $\mathbf{s} \in \mathbb{R}^{d_s}$ and $\mathbf{a} \in \mathbb{R}^{d_a}$. Let $b$ be the number of bins used to discretize each dimension of the state and action space, and flag vector $\mathbf{f}=\{f_{(1, 1, 1)}, f_{(1, 1, 2)},\cdots, f_{(d_s\cdot b, d_a \cdot b, d_s\cdot b)}\}$ where $f_{(\mathbf{i},\mathbf{j},\mathbf{k})} = 1$ if there exists data in bin index $(\mathbf{i},\mathbf{j},\mathbf{k})$, and 0 otherwise. A paralleled CDRM with $W$ weight parameters, $L$ Langevin steps, and $S$ samples is asymptotically faster than bin search-based inference when $b\cdot d_s > L\cdot W$.
\end{remark}
\vspace{-.7em}
\begin{proof}
Given input $(\mathbf{s}_t, \mathbf{a}_t)$, the bin search-based inference involves three steps: (1) calculate bin indices $\mathbf{i}, \mathbf{j}$, (2) iterate through bins for index $\mathbf{k}$ and find valid bin indices $\mathcal{K}:= \{\mathbf{k}|f_{(\mathbf{i},\mathbf{j},\mathbf{k})} = 1\}$, and (3) evaluate valid bin indices and calculate prediction and uncertainties. The first step takes $O(1)$ when $\mathcal{S}$ and $b$ are known. The second step iterates through $b\cdot d_s$ bins and takes $O(b\cdot d_s)$ time. The last step takes $O(|\mathcal{K}|)$ time, so the whole procedure takes $O(b\cdot d_s+|\mathcal{K}|)$ time.

For CDRM inference in Algorithm \ref{Alg. CDRM Inference}, we assume the number of Langevin dynamics samples is less than maximum batch size and hence the number of samples does not affect runtime. Forward pass takes $O(W)$ time and evaluating gradient (line 3, Algorithm \ref{Alg. CDRM Inference}) takes $O(W)$ time. Langevin sampling takes $L$ steps where each step requires evaluating the aforementioned gradient. The prediction and uncertainty quantification step evaluates $|\mathcal{X}|$ number of samples. Hence, CDRM inference takes $O(L\cdot W+|\mathcal{X}|)$ time.

Hence, CDRM is asymptotically faster than bin search-based inference method when $b\cdot d_s > L\cdot W$, because $|\mathcal{X}|$ would be approximately equal to $|\mathcal{K}|$. This inequality shows that CDRM would be more efficient for problems with high dimension state spaces $d_s$ such as images, requires high granularity (high $b$), or can converge with a few Langevin sampling steps. In practice, the number of weight parameters $W$ grows with the number of state dimensions $d_s$. As a result, the major advantage of CDRM compared to bin search-based inference is the capability to provide uncertainty estimation at a high granularity, because the its inference procedure samples from continuous spaces instead of discretized spaces.
\end{proof}
\vspace{-1em}
In addition to faster inference for problem that requires high granularity, CDRM has better memory complexity compared to bin search-based inference. Using the definitions in Remark \ref{Remark inference bin-search-based method}, bin search-based method requires $O(d_s^2\cdot d_a\cdot b^3)$ memory, while CDRM requires $O(W)$ memory. In practice, it is usually the case that $d_s^2\cdot d_a\cdot b^3 >> W$.

\begin{table}[]
\scriptsize
\centering
\caption{Comparison of the computational complexities of operations, where $N$ is dataset size, $S$ is sample size, $M$ is model ensemble size, and $W$ is weight parameter size.}
\label{Tab. complexity comparison}
\begin{tabular}{r|cccccc}
\toprule
           & GP (kernel learning)                      & BNN                  & MC Dropout    & Ensemble                   & CDRM  & CDRM (batch)  \\
\midrule
Memory     & $O(N^2)$                 & $O(W)$               & $O(W)$        & $O(M\cdot W)$        & $O(W)$  & $O(W)$      \\
Prediction & $O(N^2)$ (after fitting)     & $O(S\cdot W)$        & $O(S\cdot W)$ & $O(M\cdot W)$        & $O(S\cdot L\cdot W)$ & $O(L\cdot W)$ \\
Training   & $O(N^3)$ & $O(N\cdot S\cdot W)$ & $O(N\cdot W)$ & $O(N\cdot M\cdot W)$ & $O(N\cdot L\cdot W)$ & $O(N\cdot L\cdot W)$ \\
\bottomrule
\end{tabular}
\vspace{-2em}
\end{table}

Furthermore, we compare CDRM with GP (kernel learning) \cite{duvenaud2014automatic}, BNN \cite{blundell2015weight}, MC Dropout \cite{gal2016dropout}, and model ensemble methods \cite{lakshminarayanan2017simple} in terms of (i) memory complexity to store a model, (ii) computational complexity for querying predictions, and (iii) model training. The complexities are summarized in Table \ref{Tab. complexity comparison}. During prediction, CDRM requires iterating through $L$ Langevin sampling steps for $S$ samples, which results in higher time complexity compared to baselines such as BNN, which only requires $S$ forward pass outputs. CDRM can circumvent this high computational complexity by processing Langevin sampling steps in batches. Assuming $S$ is smaller than the maximum batch size, CDRM prediction only takes $O(L\cdot W)$ time, where $L$ is the number of Langevin steps and $W$ is the weight parameter size.

\section{Experiments}

\subsection{Toy Problem} \label{Sec: toy problem}

We first evaluate CDRM on two deliberately constructed toy problems to establish the correctness of CDRM's uncertainty estimates. Inspired by \cite{chan2024hyper}, we construct toy problems that contains all three uncertainty scenarios: (i) low AU, low EU (clear ground-truth prediction), (ii) high EU (no data), and (iii) high AU (high randomness data). Specifically, we generate training dataset according to the following function:
\begin{equation}
y = \begin{cases}
    \sin(x) & x \in [-1.0, -0.33) \quad\text{low AU, low EU region}\\
    \text{null} & x \in [-0.33, 0.33) \quad\text{high EU region}\\
    \sin(x) + \mathcal{N}(0, \sigma_\eta^2) & x \in [0.33, 1.0] \quad\text{high AU region}\\
\end{cases}
\end{equation}
The first dataset is directly generated from the above equation. The output distribution of any input is a unimodal distribution, we call this dataset the \textbf{unimodal} toy dataset. We also evaluate CDRM's capability to correctly estimate EU and AU in the case where the output distribution may have multiple modes that are clearly distinguishable qualitatively. Hence, we augment the unimodal dataset by appending the duplicate of the entire dataset, but change every $y$ to $-y$. After augmentation, for $x\in [-1.0, -0.33), [0.33, 1.0]$ the distribution of $y$ would potentially be multimodal distribution. We call the second dataset \textbf{multimodal} toy dataset. 

\noindent\textbf{Model training.} We train CDRM on both unimodal and multimodal datasets. The model structure is a three-layer perceptron with weight numbers $[64, 128, 64]$. During training, we used Langevin sampling steps = $10$, Langevin step size = $0.1$, Langevin noise scale = $0.01$, number of negative samples per epoch = $32$, and the training stops at $100$ epochs.

\noindent\textbf{AU estimation.} We test if CDRM correctly output high AU for regions with noise added to the $y$ values and low AU for regions where no noise is applied. For no-noise region $x \in [-1.0, -0.33)$, the AU estimates outputted by CDRM across five independent runs are $\{0.038, 0.037, 0.047, 0.038, 0.042\}$. all estimates are close to $0$, indicating a correct prediction of the lack of AU. For $x \in [0.33, 1.0]$, noise is applied, the AU estimates outputted by CDRM are $\{0.301, 0.277, 0.301, 0.279, 0.292\}$, which is sufficient for correct classification of the existence of AU.

\begin{figure}[t]
    \centering
    \includegraphics[width=\linewidth]{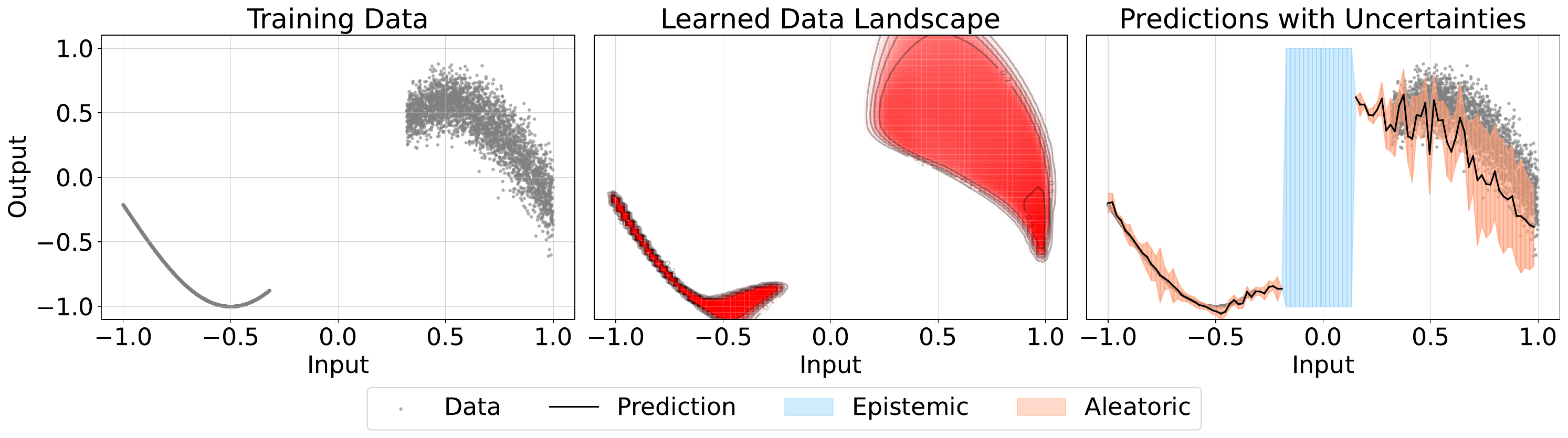}
    \vspace{-2em}
    \caption{Qualitative results of testing CDRM on a \textbf{unimodal} toy problem.}
    \vspace{-1em}
    \label{Fig. toy, single-mode}
\end{figure}

\begin{figure}[t]
    \centering
    \includegraphics[width=\linewidth]{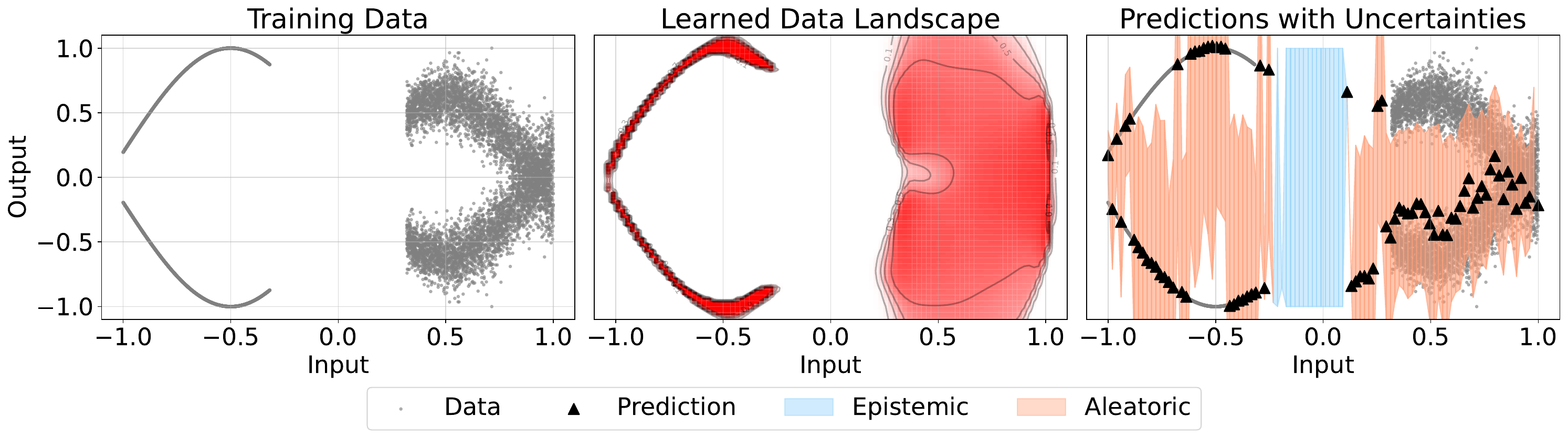}
    \vspace{-2em}
    \caption{Qualitative results of testing CDRM on a \textbf{multimodal} toy problem.}
    \vspace{-1em}
    \label{Fig. toy, multi-mode}
\end{figure}

\noindent\textbf{EU estimation.} Unlike AU, EU is dependent of the training data and hence cannot be faithfully represented by a model which has a loss function defined on a distribution over model parameters \cite{bengs2022pitfalls}. We follow the works in \cite{chan2024hyper, ekmekci2022uncertainty} and evaluate CDRM's EU estimation qualitatively. A correct EU estimation would mean that the model output $1$ for $x\in[-0.33, 0.33)$. Qualitative esults are described as follows.

\noindent\textbf{Qualitative results.} We plotted the results of running CDRM on unimodal dataset in Figure \ref{Fig. toy, single-mode} and the results on multimodal dataset in Figure \ref{Fig. toy, multi-mode}. To facilitate comparison, we plot the training data in the left most plot. To demonstrate the resulting CDRM after training, we iterate through the input space of the trained model and plotted the model output as color-mesh in the middle plot. The right most plot shows the prediction and uncertainty estimates of CDRM. We observe that, in both unimodal and multimodal datasets, CDRM outputs $EU=1$ for the no-data region in the middle, indicating that CDRM correctly estimate EU when lack data. CDRM also correctly estimated higher AU for multimodal dataset compared to unimodal dataset. 

\noindent\textbf{Baselines.} We evaluated GP \cite{duvenaud2014automatic}, BNN \cite{blundell2015weight}, MC Dropout \cite{gal2016dropout}, Deep Ensemble \cite{lakshminarayanan2017simple}, Deep Evidential Regression \cite{amini2020deep}, and HyperDM \cite{chan2024hyper} for both unimodal and multimodal datasets as baselines. Their qualitative results can be found in supplementary material by the link in abstract. All of them either do not distinguish between AU and EU or incorrectly estimate either AU or EU, as well as fail to produce a meaningful prediction when tested on multimodal dataset.

\begin{figure}
\begin{center}
    \begin{minipage}[t]{.33\linewidth}
    \includegraphics[width=\linewidth]{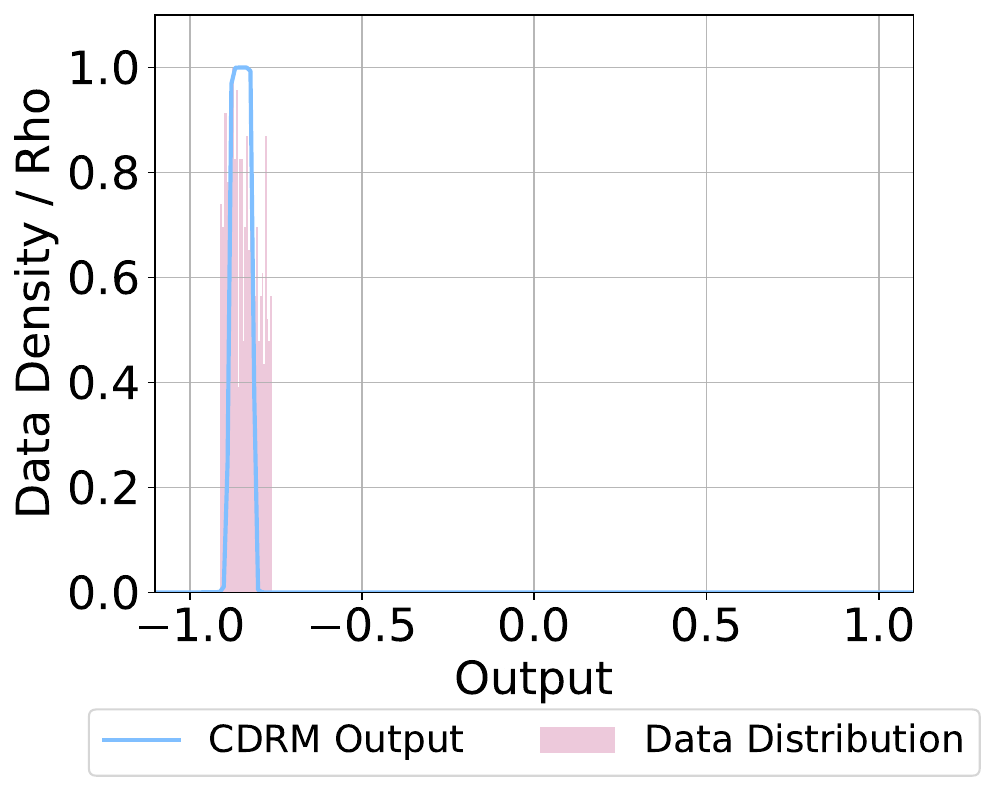}%
  \end{minipage}\hfil
  \begin{minipage}[t]{.33\linewidth}
    \includegraphics[width=\linewidth]{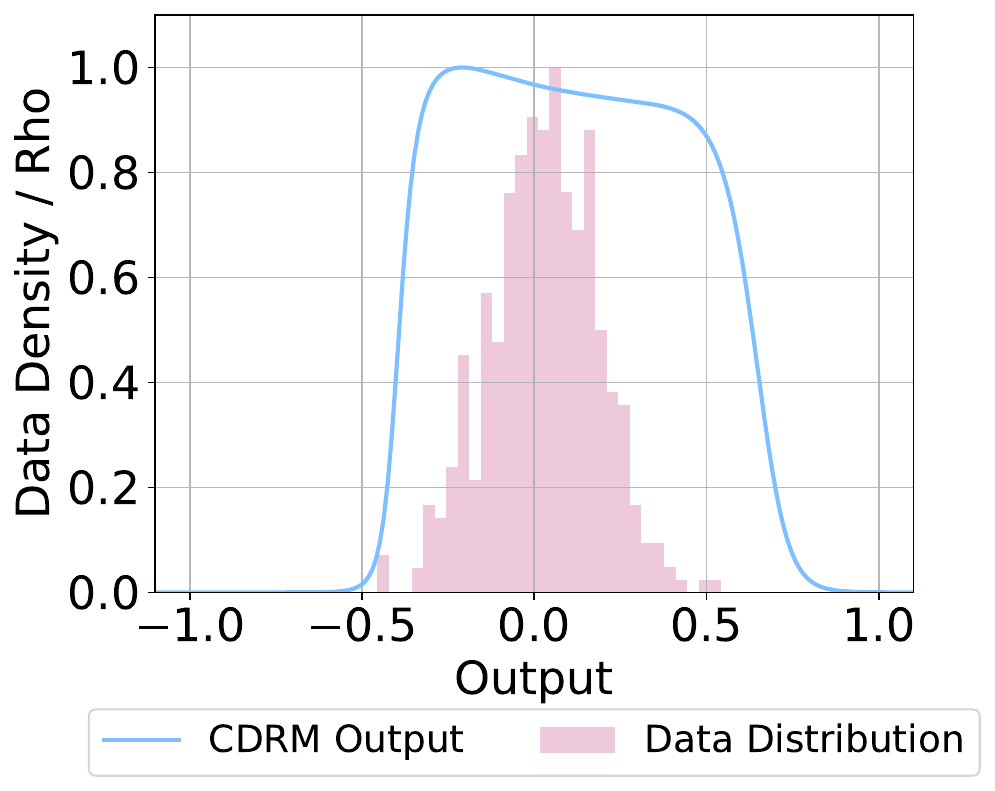}%
  \end{minipage}\hfil
  \begin{minipage}[t]{.32\linewidth}
    \includegraphics[width=\linewidth]{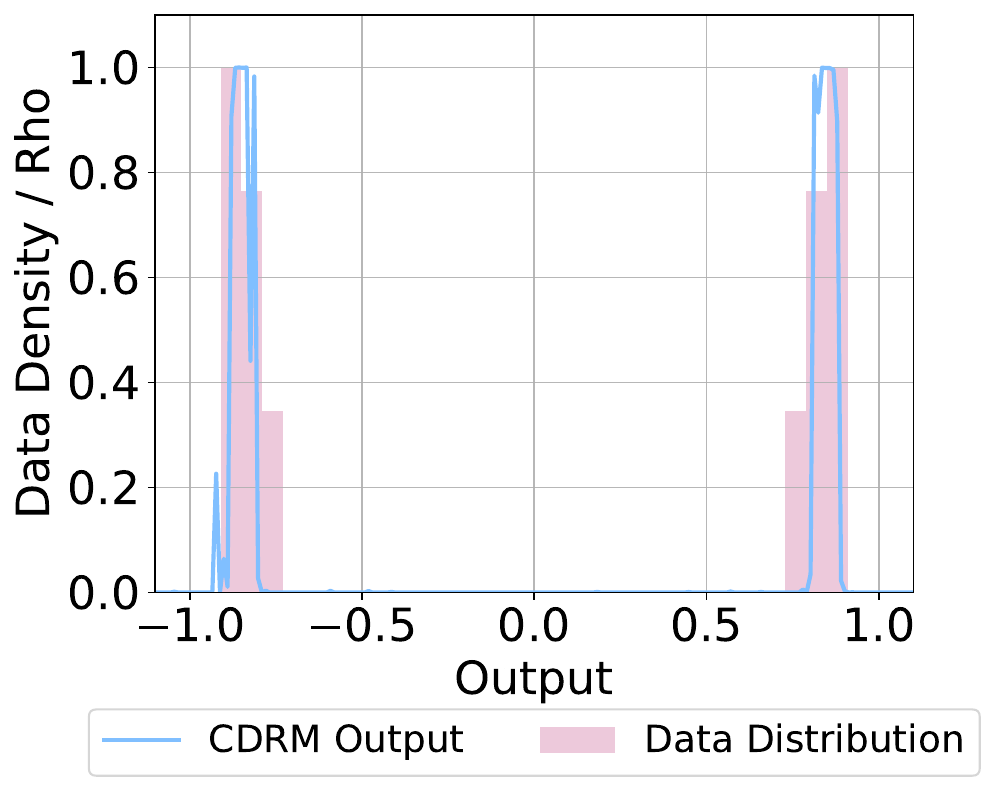}%
  \end{minipage}%
\end{center}
\vspace{-2em}
\caption{Comparison of CDRM's output and training data distribution. From left to right: $x = -0.7$ and $x = 0.9$ for unimodal dataset, and $x = -0.7$ for multimodal dataset.}
\vspace{-2em}
\label{Fig: CDRM output vs training data distribution}
\end{figure}

\noindent\textbf{Case study on CDRM's outputs.} To demonstrate the output of CDRM in-action, we compare CDRM's output and the true distribution of the training data in Figure \ref{Fig: CDRM output vs training data distribution}. We observe that CDRM correctly learns a function landscape corresponding to the presence of data. 

\subsection{Room Exploration}

\begin{wrapfigure}{r}{0.22\textwidth}
\vspace{-4em}
  \begin{center}
    \includegraphics[width=0.20\textwidth]{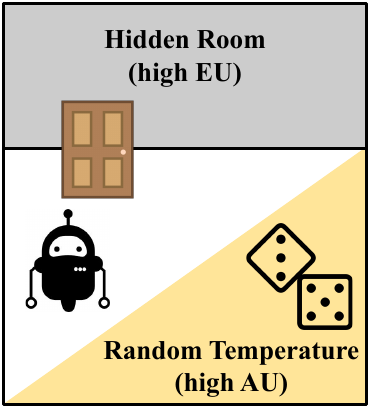}
  \end{center}
  \vspace{-2em}
  \caption{Illustration of the room exploration experiment.}
  \vspace{-1em}
  \label{Fig: 2 dimensinoal room}
\end{wrapfigure}

We designed a challenging experiment to simulate a robot exploring a two-dimensional room, illustrated in Figure \ref{Fig: 2 dimensinoal room}. The robot's position at any time can be described by a coordinate state vector. We designed two special regions in the environment: (1) a random temperature region, where the agent's sensor detects a temperature reading drawn from $\mathcal{N}(1, 0.5)$; and (2) a hidden room that the robot cannot explore. The robot randomly explores the room for $N$ time steps, recording both its coordinates and the sensed temperature $\kappa$ at each step. The resulting dataset $\mathcal{D}={(\mathbf{s}_t, \kappa_t)}$ is used to train a regressive temperature model that predicts temperature given coordinates. After training temperature models using CDRM and baseline methods, we test their AUROC and AUPRC for classifying the presence of AU and EU. For ground truth, the random temperature region has AU, the hidden room has EU, and the remaining regions have neither. The challenge lies in \textit{disentangling} AU and EU from a single dataset containing both uncertainties, as many existing methods only output a single uncertainty estimate without distinguishing between the two types.

\noindent\textbf{Baselines.} We compare CDRM with GP \cite{duvenaud2014automatic}, MC Dropout \cite{gal2016dropout}, Deep Ensemble \cite{lakshminarayanan2017simple}, Deep Evidential Regression \cite{amini2020deep}, Deep Posterior Sampling (DPS) \cite{adler2019deep}, and BNN \cite{blundell2015weight}.

\begin{table}[]
\caption{Room exploration experiment results. Best is in \textbf{bold} and the second best is in \textit{italic}.}
\vspace{-0.5em}
\label{Tab: room exploration experiment results}
\begin{tabular}{rr|ccccccc}
\toprule
                           &       & GP     & MC Dropout & Ensemble & DER    & DPS    & BNN    & CDRM \\
\midrule
\multirow{2}{*}{AU} & AUPRC $\uparrow$ & 0.1971 & 0.3368     & 0.2921   & \textbf{0.7673} & 0.1651 & 0.3413 &   \textit{0.7170}  \\
                           & AUROC $\uparrow$ & 0.0596 & 0.5893     & 0.4966   & \textit{0.7902} & 0.4230 & 0.5886 &   \textbf{0.8876}  \\
\midrule
\multirow{2}{*}{EU} & AUPRC $\uparrow$ & 0.5539 & \textit{0.8396}     & 0.3925   & 0.6801 & 0.7687 & 0.8217 &  \textbf{0.9961}  \\
                           & AUROC $\uparrow$ & 0.7692 & \textit{0.9056}     & 0.6243   & 0.7034 & 0.6537 & 0.8962 &  \textbf{0.9981} \\
\bottomrule
\end{tabular}
\vspace{-1.5em}
\end{table}

\noindent\textbf{Quantitative results.} The experimental results are shown in Table \ref{Tab: room exploration experiment results}. We observed that many methods excel at estimating only one type of uncertainty and exhibit major drawbacks in correctly classifying the other type. MC Dropout, DPS, and BNN excel at classifying EU but have poor performance in recognizing AU. DER achieved a balanced performance in classifying AU and EU, showing a major advantage in classifying AU compared to other baseline methods. Notably, CDRM is the only method that excels at both AU and EU classification, demonstrating its superior performance in disentangling the two types of uncertainties.





\section{Limitations and Future Work} \label{Sec. future work}
We note two key limitations found during experiments. First, CDRM has high computational overhead during training since achieving zero output for non-data distribution inputs in high-dimensional spaces requires many Langevin samples and iterations. Second, CDRM captures only the existence of data but not its relative abundance. While we complemented CDRM with KDE to address this, modifying the training procedure might enable learning both properties simultaneously. For future work, we note that our naive method of generating single predictions for multimodal next-state distributions, though avoiding averaging problems, fails to reflect CDRM's full belief. Multimodal next-state prediction using CDRM could enable more informative predictions for control and reinforcement learning applications.

\section{Conclusion}

We studied aleatoric and epistemic uncertainty estimation in system dynamics models using transition data. We proposed the Compressed Data Representation Model (CDRM), a framework that learns data distributions using neural networks. Our Langevin dynamics sampling inference procedure enables effective uncertainty estimation from CDRM. Experiments demonstrated CDRM's superior ability to disentangle aleatoric and epistemic uncertainties in datasets containing both types. CDRM also effectively handles multimodal output distributions, a challenging scenario where existing methods fail.

\acks{This work was supported in part by the NSF Grant \#2239458, the UC National Laboratory Fees Research Program grant \#69763, and an UC Merced Fall 2023 Climate Action Seed Competition Grant. Any opinions, findings, and conclusions expressed in this material are those of the authors and do not necessarily reflect the views of the funding agencies.}

\bibliography{ref}





\end{document}